\pdfoutput=1
\documentclass[11pt]{article}

\usepackage{EMNLP2022}

\usepackage{times}
\usepackage{latexsym}

\usepackage[T1]{fontenc}
\usepackage[utf8]{inputenc}

\usepackage{microtype}

\usepackage{enumitem}
\setenumerate[1]{itemsep=0pt,partopsep=0pt,parsep=\parskip,topsep2pt}
\setitemize[1]{itemsep=0pt,partopsep=0pt,parsep=\parskip,topsep=2pt}
\setdescription{itemsep=0pt,partopsep=0pt,parsep=\parskip,topsep=2pt}

\usepackage{inconsolata}

\usepackage{booktabs}
\usepackage{hyperref}
\usepackage{url}
\usepackage{graphicx}
\usepackage{xcolor}
\usepackage{amsmath}
\usepackage{amssymb}
\usepackage{color, colortbl}
\usepackage{makecell}
\usepackage{microtype}
\usepackage{xspace}
\usepackage{ulem}
\usepackage{array, tabularx, caption, boldline}
\usepackage{cellspace}
\usepackage{algpseudocode}  
\usepackage{amsmath}
\usepackage{multicol}  
\usepackage{multirow} 
\usepackage{flexisym}
\usepackage{graphicx}  %Required
\usepackage{array}
\usepackage{multicol}
\usepackage{multirow}
\usepackage{amsmath}
\usepackage{booktabs}
\usepackage{array}
\usepackage{soul}
\usepackage{vcell}
\usepackage{pbox}
\usepackage{enumitem}
\usepackage{graphics}
\usepackage{wrapfig}
\usepackage{makecell}
\newcommand{\tabincell}[2]{\begin{tabular}{@{}#1@{}}#2\end{tabular}}

\usepackage{tabularx}
\makeatletter
\def\hlinewd#1{%
\noalign{\ifnum0=`}\fi\hrule \@height #1 %
\futurelet\reserved@a\@xhline}
\makeatother

\definecolor{darkgreen}{rgb}{0, 0.25, 0.}
\definecolor{darkblue}{rgb}{0, 0., 0.35}
\definecolor{darkorange}{rgb}{255, 140, 0}
\definecolor{LightCyan}{rgb}{0.88,1,1}

\title{Controllable Dialogue Simulation with In-Context Learning}

\author{Zekun Li\textsuperscript{1}, Wenhu Chen\textsuperscript{2}, Shiyang Li\textsuperscript{1}, Hong Wang\textsuperscript{1}, Jing Qian\textsuperscript{1}, Xifeng Yan\textsuperscript{1} \\
        \textsuperscript{1}University of California, Santa Barbara\\
        \textsuperscript{2}University of Waterloo, Vector Institute\\
        \tt \{zekunli, shiyangli, hongwang600, jing\_qian, xyan\}@cs.ucsb.edu\\
        \tt wenhuchen@uwaterloo.ca
}

\begin{document}
\maketitle

\begin{abstract}
Building dialogue systems requires a large corpus of annotated dialogues. Such datasets are usually created via crowdsourcing, which is expensive and time-consuming. In this paper, we propose \textsc{Dialogic}\footnote{\textsc{\textbf{DIALOG}ue simulation with \textbf{I}n-\textbf{C}ontext learning}}, a novel dialogue simulation method based on large language model in-context learning to automate dataset creation. Seeded with a few annotated dialogues, \textsc{Dialogic} automatically selects in-context examples for demonstration and prompts GPT-3 to generate new dialogues and annotations in a controllable way. Our method can rapidly expand a small set of dialogue data with minimum or zero \textit{human involvement} and \textit{parameter update} and is thus much more cost-efficient and time-saving than crowdsourcing. Experimental results on the MultiWOZ dataset demonstrate that training a model on the simulated dialogues leads to even better performance than using the same amount of human-generated dialogues under the challenging low-resource settings, with as few as 85 dialogues as a seed. When enough data is available, our method can still serve as an effective data augmentation method. Human evaluation results also show that our simulated dialogues have near-human fluency and annotation accuracy. The code and data are available at \textbf{\url{https://github.com/Leezekun/dialogic}}.
\end{abstract}

\begin{figure*}[ht!]
    \centering
    \includegraphics[width = 1\linewidth]{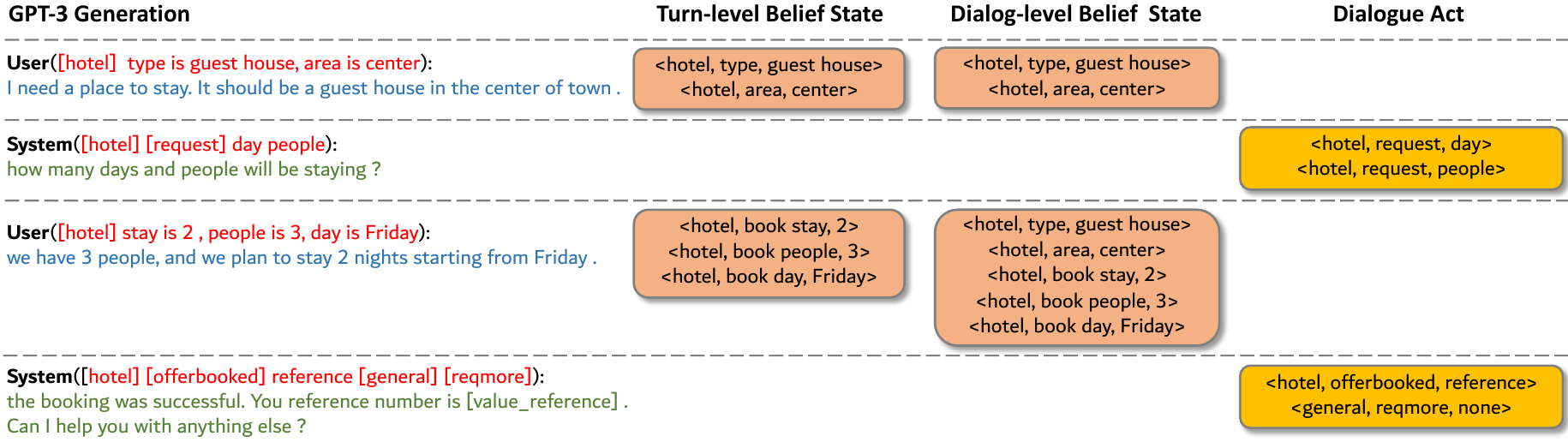}
    % \caption{Illustration of part of an annotated dialogue generated by our method. All the left part are generated by GPT-3. The \textcolor{darkblue}{user utterances} in blue color and \textcolor{darkgreen}{the system responses} in green color. The red part in the brackets can be parsed to annotations, i.e., belief states and dialog acts in the right part, which can also help us track and control the generation process and dialogue flow. A complete dialogue generated by GPT-3 is shown in Appendix~\ref{sect:appendix2}.
    \caption{Illustration of a part of an annotated dialogue generated by our method. \textbf{Left}: the conversations and annotations are generated simultaneously by GPT-3, where the \textcolor{darkblue}{user utterances} are in blue, the \textcolor{darkgreen}{system responses} are in green, and the annotations are in red. \textbf{Right}: the structured annotation obtained by parsing the GPT-3's generation shown on the left. Best viewed in color. A complete generated dialogue is shown in Appendix~\ref{sect:dst_example} as Table~\ref{tab:whole_generation_process}.
    }
    \label{fig:illustration}
    \vspace{-2mm}
\end{figure*}

\section{Introduction}
Task-oriented dialogue (TOD) systems can assist users in completing tasks such as booking a restaurant or making an appointment. Building such a dialogue system requires a large corpus of annotated dialogues~\citep{Wu2020TODBERTPN}, which is costly to obtain in terms of money and time.

One popular approach to collecting and annotating task-oriented dialogues is crowdsourcing via a Wizard-of-Oz setup~\citep{woz,kvret,multiwoz}, where crowdworkers produce conversations. Significant annotation efforts are further needed to label intent, entities, etc. Prior work has been proposed to minimize the cost and effort in data collection by hiring crowdworkers or leveraging user simulators to interact with existing dialogue systems~\citep{dstc,shah2018building,shah2018bootstrapping,Papangelis2019CollaborativeMD,zhao2019rethinking,sgd,tseng2021transferable}.
However, the dependency on existing dialogue systems leave the developers with a classic chicken-and-egg problem. In addition, developing such user simulators typically requires considerable handcrafting and human involvement. 
% Typically they utilize pre-defined rules and hand-crafted templates, which can not guarantee the naturalness and diversity of generated dialogues and require considerable engineering efforts and human involvement. 
% To solve the problem, \citet{Mohapatra2020SimulatedCF} proposed a method that uses GPT-2~\citep{gpt2} to simulate the user, system, and their interactions. However, it requires human-written instructions to evoke the generation and a considerable amount of seed dialogues for simulator training to ensure acceptable generation quality.
% However, it requires human-written instructions to evoke the simulation and a considerable amount of seed dialogues for the train the simulator.

In recent years, large language models (LLMs)~\citep{gpt3,lieber2021jurassic,rae2021scaling,lamda,megatron-turing} demonstrate strong in-context learning capability. Provided with a few in-context examples, the LLMs, such as GPT-3~\citep{gpt3}, can generate text with similar patterns without fine-tuning. This capability has been leveraged to synthesize training data in a few NLP tasks~\citep{wang2021towards,liu2022wanli}. Although there have been methods that generate training data for a single component in the TOD systems~\citep{limaking}, there hasn't been a plausible solution to generate whole dialogues with annotations for end-to-end training due to its complex nature of involving multi-turn interactions, multiple possible logic flows, and multiple types of annotations.

To address the challenge, we introduce a controllable dialogue simulation method \textsc{Dialogic} for dialogue dataset creation. Seeded with a few seed dialogues, \textsc{Dialogic} automatically selects in-context examples for demonstration and prompts LLMs such as GPT-3 to generate annotated dialogues in a controllable way. \textsc{Dialogic} can play the roles of both user and system simulator. Figure~\ref{fig:illustration} illustrates a partial example. For the user side, GPT-3 is prompted first to generate the turn-level user goal (belief state), conditioned on which the user utterance that expresses the goal will be generated. Likewise, we prompt GPT-3 to generate the dialog act for the system side and then the corresponding system response. We also propose automatic verification and revision methods to mitigate annotation errors. 
% The generated belief states and dialog acts are used as annotations and can also help us control the generation process. 

This paper has two key insights. First, leveraging the in-context learning ability of LLMs, our method can simulate both the user and system side to generate annotated dialogues by learning from a few examples. Except for the minimal efforts in collecting the small seed dataset and training an auxiliary model on that, the simulation process is free of \textit{human involvement} and \textit{parameter update}, making our method much cheaper and faster than crowdsourcing in dataset creation. Specifically, a large-scale and high-quality dataset such as MultiWOZ~\cite{multiwoz} can be created using our method within only several hours. Second, we design controllable dialogue generation strategies to overcome the deficiency of GPT-3 in lack of reliability and interpretability. We also investigate effective representations and selection strategies of in-context dialogue examples for LLMs to better leverage their in-context learning capabilities.

We conduct experiments on MultiWOZ2.3~\citep{multiwoz23} dataset. Remarkably, in the challenging low resource settings where as low as only 85 seed dialogues (1\% of the whole training dataset) are given, the dialogues simulated by our method lead to even better model performance than the same amount of human-generated dialogues. 
\textsc{Dialogic} can also serve as an effective data augmentation method when the full training set is provided. Human evaluations indicate that our simulated dialogues have comparable fluency, annotation accuracy, and more diverse dialogue flows than human-generated dialogues. Our results demonstrate the promise of leveraging large language model to automate the complex dialogue dataset creation. We have released the code and simulated data to facilitate future studies.\footnote{\href{https://github.com/Leezekun/dialogic}{https://github.com/Leezekun/dialogic}}

\begin{figure*}[ht!]
    \centering
    \includegraphics[width = 1.0\linewidth]{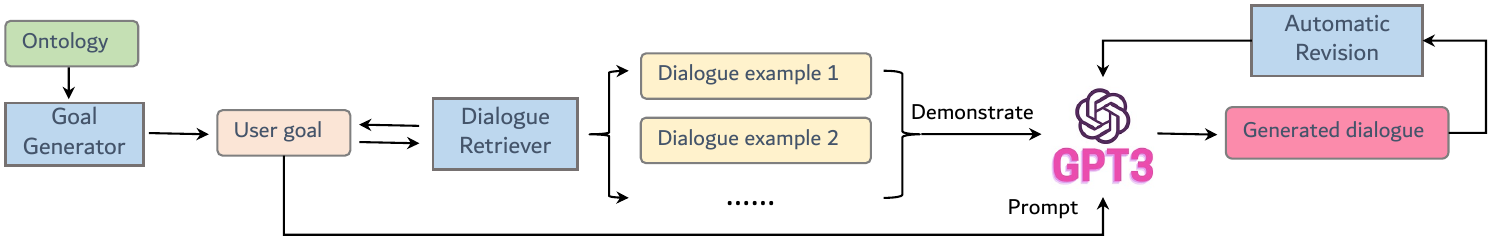}
    \caption{Overview of the proposed method.}
    \label{fig:overview}
    % \vspace{-2mm}
\end{figure*}

\section{Related Work}
\subsection{Dialogue Collection and Simulation}
Building end-to-end dialogue systems heavily relies on annotated training data. Wizard-of-Oz~\citep{kelley1984iterative}, as a popular approach, is able to produce high-quality conversations but totally relies on human efforts~\citep{woz,kvret,frames,multiwoz}. There are also dialogue corpora of interactions between humans and existing dialogue systems or APIs~\citep{dstc,williams2014dialog,raux2005let}. To further reduce human efforts, user simulators are leveraged to interact with the system via reinforcement learning or self-play~\citep{shah2018building,shah2018bootstrapping,Papangelis2019CollaborativeMD,zhao2019rethinking,sgd,tseng2021transferable}. However, existing dialogue systems or APIs are still needed, which restricts these solutions to existing domains. To this end, \citet{Mohapatra2020SimulatedCF} proposed a method that utilizes GPT-2~\citep{gpt2} to simulate both the user and system side. However, this method still needs many dialogues to train the simulators and cannot guarantee the simulation quality in low-resource settings.

\subsection{Task-oriented Dialogue}
A task-oriented dialogue system usually consists of three components: natural language understanding (NLU) for dialogue state tracking, dialogue management (DM) for predicting the dialog act based on the dialogue states, and natural language generation (NLG) for mapping dialog act to natural language response. The annotated data of belief states, dialog acts, and system responses are needed to train these components whether in a separate way~\citep{trade,sumbt,trippy}, or an end-to-end fashion~\citep{soloist,simpletod,mintl,ubar,pptod}. In this paper, we aim to generate dialogues and their complete set of annotations.

\subsection{In-Context Learning}
As an alternative to finetuning, in-context learning with LLMs, such as GPT-3~\citep{gpt3}, can perform a new task by learning from a few in-context examples without training model parameters. Due to the superior few-shot performance and scalability, in-context learning has been applied to a wide range of NLP tasks. As for dialogue tasks, in-context learning has been increasingly deployed in tasks such as intent classification~\citep{yu2021few}, semantic parsing~\citep{shin2021few}, and dialogue state tracking~\citep{hu2022context}. \citet{madotto2021few} built an end-to-end dialogue system solely based on in-context learning. Despite its success, GPT-3 requires a large number of resources to be deployed. And its public API is charged based on the length of input text. What's worse, the limitation of input length restricts the number of in-context examples and thus the generation performance. Consequently, a few methods have been proposed to leverage GPT-3 to synthesize data to train smaller models for inference~\citep{wang2021want,wang2021towards,liu2022wanli,LI2022ExplanationsFL}. Although it is especially desirable for dialogue tasks as the input prompt of dialogues is usually lengthy, there hasn't been a plausible solution to generating annotated dialogues for developing TOD systems due to its complex nature of involving multi-turn interactions and multiple types of annotations.

\section{Method}
In this paper, we introduce a novel method \textsc{Dialogic} to simulate annotated dialogues for building task-oriented dialogue systems based on language model in-context learning. The only requirements are a small seed dataset $\mathcal{D}_{s}$ consisting of a few annotated dialogues and an ontology $\mathcal{O}$ that includes all slots and possible slot values for each domain. 
% For example, the \textit{restaurant} domain may have slots of \textit{pricerange}, \textit{area}, \textit{food} and so on. Each slot has some possible values (e.g. \textit{pricerange} may be \textit{cheap}, \textit{moderate}, or \textit{expensive}). 
An auxiliary TOD model $\mathcal{M}$ such as SimpleTOD~\citep{simpletod} and PPTOD~\citep{pptod} trained on $\mathcal{D}_{s}$ will be used to verify and revise generated annotations. Our goal is to expand $\mathcal{D}_{s}$ by generating new dialogues. For each turn of the dialogues, we need to generate the user utterance $U$, belief state $B$, database (DB) query result $Q$, dialog act $A$, and system responses $S$ (we omit the turn index for brevity).

We will elaborate the design of our method using a well-studied task-oriented dialogue dataset MultiWOZ~\citep{multiwoz,multiwoz21,multiwoz23}, which covers 7 domains such as \textit{hotel} and \textit{restaurant}, and 24 slots such as \textit{hotel-area} and \textit{restaurant-food} (see Appendix~\ref{sect:implementation} for more details). 
To simulate the low-resource environment, we use 1\%, 5\%, 10\% of the training dataset as the seed dataset $\mathcal{D}_{s}$.

\subsection{Overview}
A partial example of a simulated dialogue is shown in Figure~\ref{fig:illustration}. The pipeline of our method is illustrated in Figure~\ref{fig:overview}. For a domain, the goal generator will take the ontology $\mathcal{O}$ as input to generate a new user goal $\mathcal{G}_i$. Then we select a few seed dialogues with similar user goals from $\mathcal{D}_{s}$ as the in-context example for GPT-3. Given the user goal $\mathcal{G}_i$ and the selected in-context examples, we leverage GPT-3 to generate a new dialogue $\mathcal{C}_i$. As the generated data may fail to satisfy our requirement, we design methods for automatic verification and revision.

\subsection{In-context Example}\label{sect:example}
\noindent\textbf{User Goal.~}
A task-oriented dialogue is a conversation where the dialogue system helps accomplish the user's goal. For a new dialogue $\mathcal{C}_i$, we first generate its user goal $\mathcal{G}_i$ based on the ontology. The user goal and belief state are a set of domain-slot-value triplets: (\textit{domain, slot\_name, slot\_value}). For example, when a user wants to book a 4-star hotel for 2 nights, and a cheap restaurant that serves Chinese food, his user goal will be \{(\textit{hotel, stars, 4}), (\textit{hotel, book stay, 2}), (\textit{restaurant, pricerange, cheap}), (\textit{restaurant, food, chinese})\}. We investigate several ways to generate the user goal, i.e., determining the domains, slots, and slot values to be selected, which will be discussed as follows.

\noindent\textbf{Example Selection.~}
Given the target user goal $\mathcal{G}_t$, we select a few seed dialogues as in-context examples, from which GPT-3 can learn to generate the target dialogue $\mathcal{C}_i$.
To achieve that, the selected dialogue examples should contain as much ontology information needed in the target dialogue (i.e., mentioned slots) as possible so that GPT-3 can mimic the ``in-domain'' generation. To measure how two dialogue goals $\mathcal{G}_i$ and $\mathcal{G}_j$ overlap, we calculate their similarity as:
\begin{equation}
w_{ij} = \left | \frac{D(\mathcal{G}_i)\bigcap D(\mathcal{G}_j)}{D(\mathcal{G}_i)\bigcup D(\mathcal{G}_j)} \right | \cdot \left | \frac{S(\mathcal{G}_i)\bigcap S(\mathcal{G}_j)}{S(\mathcal{G}_i)\bigcup S(\mathcal{G}_j)} \right | ,
\end{equation}
where $D(\mathcal{G}_i)$ and $S(\mathcal{G}_i)$ denote the set of domains and slots in the user goal $\mathcal{G}_i$, respectively. The first part is the Jaccard similarity~\citep{jaccard} of the domain set, while the second part is that of the slot set.
The probability of a dialogue $\mathcal{C}_j$ from the seed dataset $\mathcal{D}_{s}$ being sampled as in-context examples for the target dialogue $\mathcal{C}_i$ is:
\begin{equation}\label{eq:sampling}
    p_j = \frac{e^{w_{ij}/\tau}}{\sum_{\mathcal{C}_k\in\mathcal{D}_s}e^{w_{ik}/\tau}},
\end{equation}
where $\tau$ is the temperature. A higher temperature will introduce more randomness and diversity in example selection.

We investigate several ways to generate user goals and select in-context examples:
\begin{itemize}
    \item \textit{Random Sampling}: we randomly select domains, slots, and slot values to form a user goal and sample in-context examples as described in Equation~\ref{eq:sampling}. In this way, we can generate any unseen user goal and thus the corresponding dialogues. 
    However, as the number of seed dialogues is limited, it is hard to guarantee that the sampled dialogue examples can cover all the information required for generating the target dialogue. 
    \item \textit{Value Substitution}: we only substitute the slot values of the seed dialogues' user goal to form a new user goal. This method can ensure that all the required slots are mentioned in the in-context examples. However, GPT-3 will tend to replicate the in-context examples, and thus few diversity can be introduced.
    \item \textit{Combination}: we first select a few dialogues from the seed dataset and then combine their user goals to create a new goal. As the new user goal might involve too many domains and slots, we randomly drop some slots. This method can ensure that all the mentioned slots in the target user goal are covered in the examples and encourage the GPT-3 to generate diverse data. 
\end{itemize}
We experimentally found the \textit{Combination} method yields the best performance. More details, comparison, and discussion of different goal generation methods can be found in Appendix~\ref{sect:augmentation_method}.

\begin{figure}[!t]
%\scriptsi
\scriptsize
\fontsize{7.6pt}{9pt}\selectfont
\noindent\fbox{
    \parbox{.975\linewidth}{
        \textbf{Instruction}:
        You are going to book a hotel, and your requirements for the hotel are (\textcolor{red}{[hotel] type is hotel , pricerange is cheap , parking is yes , stay is 2 , day is tuesday , people is 6}). Make sure you get the booking information once booked. 
    }
}
\noindent\fbox{%
    \parbox{.99\linewidth}{%
        \textbf{Conversation}: \\ 
        \textbf{User} (\textcolor{red}{[hotel] type is hotel , pricerange is cheap}): \textcolor{darkblue}{I am looking for a place to to stay that has cheap price range it should be in a type of hotel .} \\
        \textbf{Assistant} (\textcolor{red}{[hotel] [request] area}): \textcolor{darkgreen}{okay , do you have a specific area you want to stay in ?} \\
        \textbf{User} (\textcolor{red}{[hotel] parking is yes , pricerange is cheap}): \textcolor{darkblue}{no , i just need to make sure it is cheap . oh , and i need parking .} \\
        \textbf{Assistant} (\textcolor{red}{[hotel] [inform] price choice parking type [offerbook]}): \textcolor{darkgreen}{i found [value\_choice] [value\_price] [value\_type] for you that include -s parking . do you like me to book it ?} \\
        \textbf{User} (\textcolor{red}{[hotel] stay is 3 , day is tuesday , people is 6}): \textcolor{darkblue}{yes , please . 6 people 3 nights starting on tuesday .} \\
        \textbf{Assistant} (\textcolor{red}{[hotel] [nobook] day [request] stay day}): \textcolor{darkgreen}{i am sorry but i was n't able to book that for you for [value\_day] . is there another day you would like to stay or perhaps a shorter stay ?} \\
        \textbf{User} (\textcolor{red}{[hotel] stay is 2}): \textcolor{darkblue}{how about only 2 nights .} \\
        \textbf{Assistant} (\textcolor{red}{[hotel] [offerbooked] reference [general] [reqmore]}): \textcolor{darkgreen}{booking was successful . reference number is : [value\_reference] . anything else i can do for you ?} \\
        \textbf{User} (\textcolor{red}{[general]}): \textcolor{darkblue}{no , that will be all . goodbye .} \\
        \textbf{Assistant} (\textcolor{red}{[general] [bye]}): \textcolor{darkgreen}{thank you for using our services .}
    }%
}
\caption{Illustration of an in-context example from the MultiWOZ dataset. The \textcolor{red}{user goal}, \textcolor{red}{belief states}, and \textcolor{red}{dialog acts} are in red. \textcolor{darkblue}{User utterances} are in blue, while \textcolor{darkgreen}{system responses} are in green. Best viewed in color.
}
\label{fig:examples}
\vspace{-4mm}
\end{figure}

\noindent\textbf{Demonstration.~}
To better demonstrate the desired pattern of generated data for a dialogue to GPT-3, we design the format for the example dialogues as shown in Figure~\ref{fig:examples}. The user goal and belief state are converted from a sequence of triplets to the natural language via a template. For example, the user goal of \{(\textit{hotel, stars, 4}), (\textit{hotel, book stay, 2}), (\textit{restaurant, pricerange, cheap}), (\textit{restaurant, food, chinese})\} will be converted to \textit{[hotel] star is 4 , book stay is 2 [restaurant] pricerange is cheap , food is chinese}, where \textit{[domain]} separates domains and the comma separates slots in each domain.
% Note that the MultiWOZ dataset provides user goal descriptions. However, we use the template user goal description as the \textit{instruction} because it is more concise and is in the same pattern as the belief state. 

As for the \textit{conversation} part, the desired annotations are incorporated with the utterances for each turn. For the user side, GPT-3 will generate the user utterance and the turn-level belief state, i.e., the user goal mentioned in this turn. Dialog acts and their corresponding system response are needed for the system side. Similarly, the dialog act is also a set of triplets (\textit{domain, action\_type, slot\_name}) and is converted to natural language similarly to belief states. An actual example of the demonstration can be seen in Table~\ref{tab:whole_prompt}.

\begin{figure}[!t]
    \centering
    \includegraphics[width = 1\linewidth]{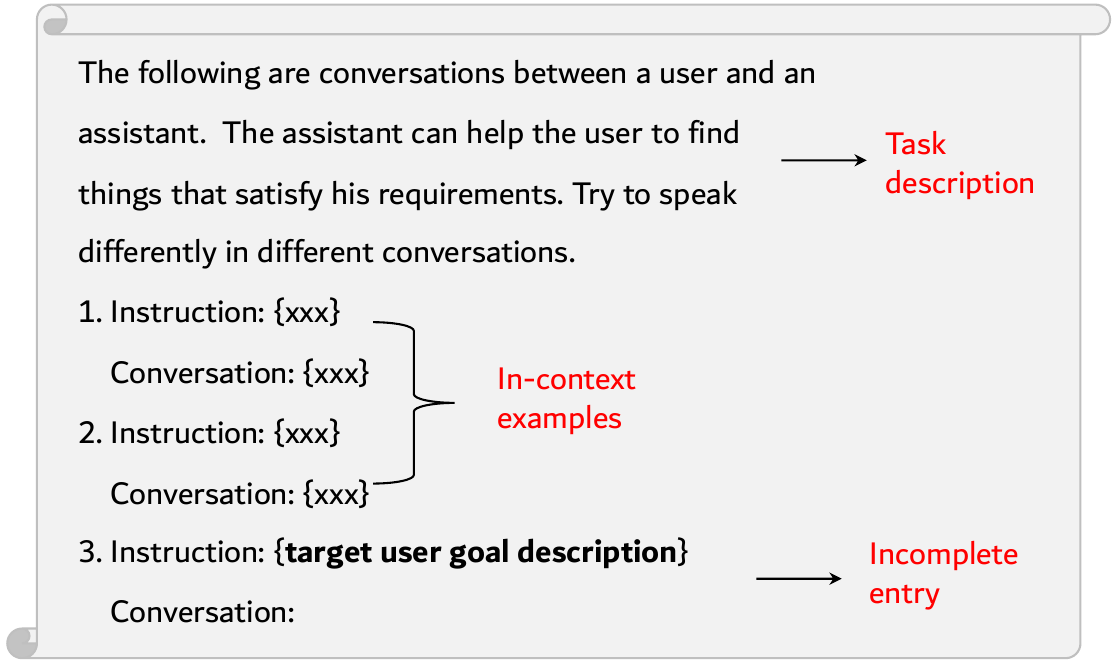}
    \caption{Template for the prompt of GPT-3 to generate new dialogues. An actual example of the complete prompt is shown in Appendix~\ref{sect:dialog_example} as Table~\ref{tab:whole_prompt}.}
    \label{fig:prompt}
    \vspace{-4mm}
\end{figure}

\subsection{Prompt Design}
Given a prompt consisting of the task description, a few in-context examples, and an incomplete entry, we instruct GPT-3 to generate text to complete the entry. A template of our prompt is shown in Figure~\ref{fig:prompt}. The format of in-context examples is described in Section~\ref{sect:example}, which consists of an \textit{instruction} (user goal), and the \textit{conversation}. As for the incomplete entry, the target user goal description is used as the \textit{instruction}, and GPT-3 will generate the corresponding \textit{conversation} in a controllable way, which will be described in the next section.

\begin{figure}[!ht]
    \centering
    \includegraphics[width = 1\linewidth]{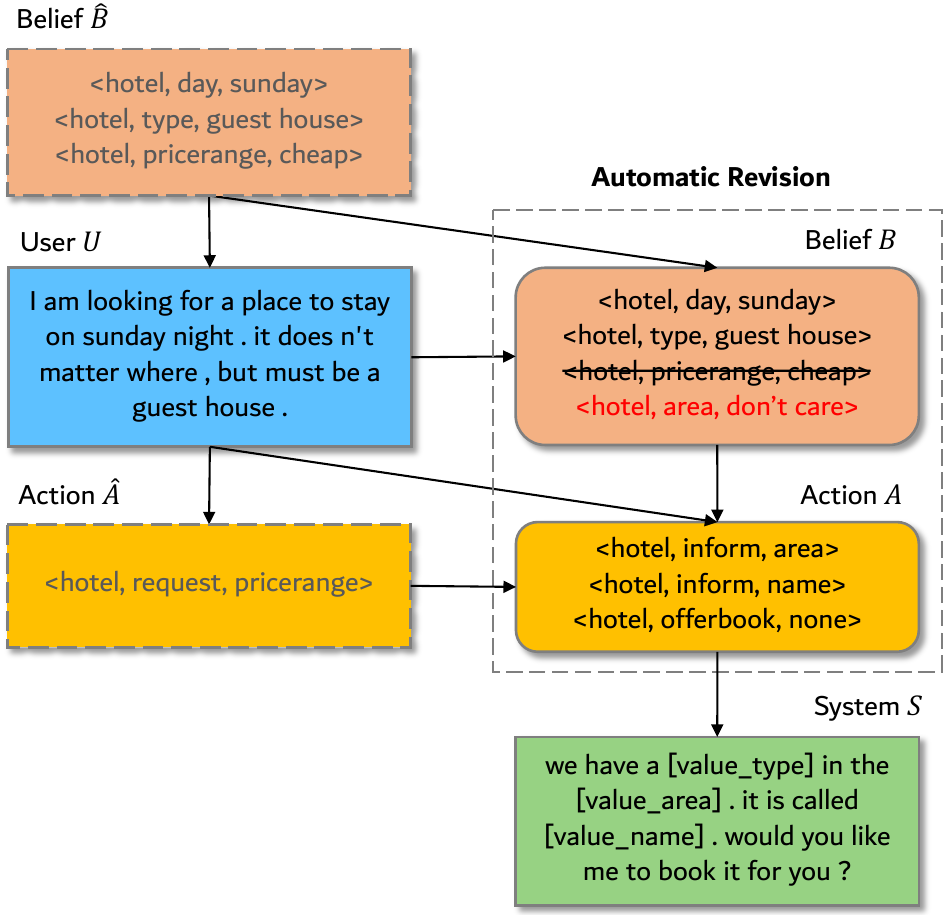}
    \caption{Illustration of the controllable generation process of a dialogue turn. An example of the generation process of a complete dialogue is shown in Appendix~\ref{sect:dialog_example} as Table~\ref{tab:whole_generation_process}.}
    \label{fig:process}
    \vspace{-4mm}
\end{figure}

\begin{figure*}[!t]
    \centering
    \includegraphics[width = 1\linewidth]{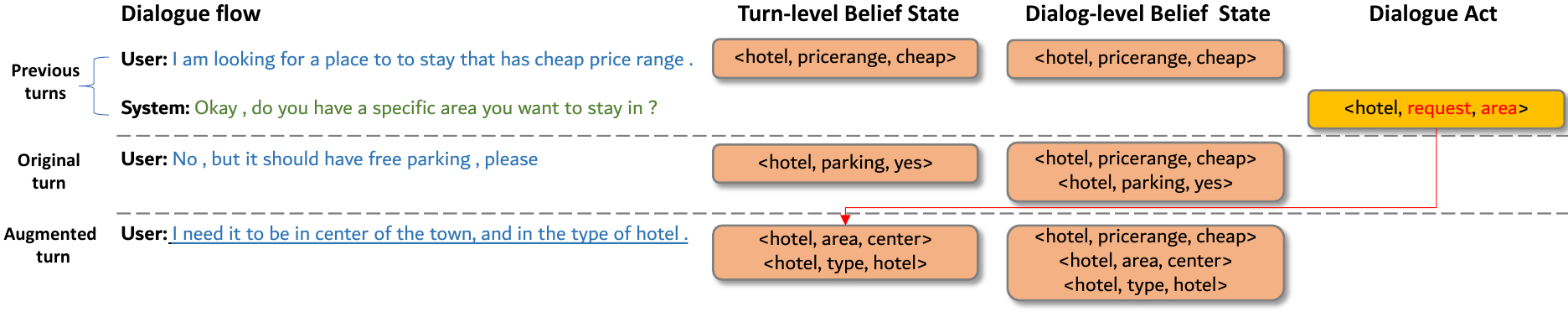}
    \caption{Illustration of the turn-level generation for DST augmentation. The turn-level belief state is decided on the dialog act of the last turn. The user utterance, which is underlined, is generated conditioned on the turn-level belief state. The newly generated user turn will be concatenated with previous turns of the original dialogue to form a new dialogue. An example of the generated user turn is shown in Appendix~\ref{sect:dst_example} as Table~\ref{tab:dst_prompt}.}
    \label{fig:dst}
    \vspace{-4mm}
\end{figure*}

\subsection{Controllable Dialogue Generation}\label{sect:generation}
Considering GPT-3's known deficiencies in lack of reliability and interpretability, we propose methods to control GPT-3 when it generates dialogue data. In addition, we design automatic revision methods to minimize potential annotation errors. Figure~\ref{fig:process} illustrates the controllable generation process of a dialogue turn.

For the user side, GPT-3 will generate the belief state $\hat{B}$ and the corresponding user utterance $U$. The belief state is expected to be consistent with the the user utterance. We keep $U$ unchanged as the final user utterance and check the annotation errors in the generated belief state $\hat{B}$, which can be categorized into two types. Taking the example in Figure~\ref{fig:process} for illustration, \textit{(hotel, stay, 1)}, as a part of the original generated belief $\hat{B}$, doesn't appear in the user utterance, which is called \textit{over-generation}. On the contrary, the value \textit{don't care} for slot \textit{hotel-area} is mentioned by the user but not included in $U$, which is called \textit{de-generation}~\citep{li2020coco}. We utilize an \textit{auxiliary generator} and \textit{slot-value match filter} to mitigate \textit{de-generation} and \textit{over-generation} issues, respectively.

\noindent\textbf{Auxiliary Generator.~} To tackle the \textit{de-generation} issue, we try to detect as many mentioned slots in the user utterances as possible. 
% To achieve that, we utilize an auxiliary TOD model $\mathcal{M}$ trained on the seed dataset $\mathcal{D}_{s}$ to generate its predicted belief state:
% \begin{equation}\label{eq:verifier}
%     \bar{B} = \mathcal{M}(X, U),
% \end{equation}
% where $X$ denotes the dialogue context consisting of the user and system utterance s of all previous turns. 
To this end, we utilize an auxiliary TOD model $\mathcal{M}$ trained on the seed dataset $\mathcal{D}_{s}$ to generate its predicted belief state $\bar{B}$, conditioned on the dialogue context of all previous turns and user utterance $U$ of the current turn. $\bar{B}$ could be complementary to $\hat{B}$ generated by GPT-3. 
% With increasing size of seed data, the auxiliary generator can better generate the correct slots not detected by GPT-3. Even when the seed data is limited, the auxiliary generator can still help to mitigate the de-generation issue, as we found that GPT-3 sometimes only generate the user utterance and forgets to generate any belief state.
We found that GPT-3 sometimes forgets to generate all or even any belief state. If not corrected, GPT-3 will continue the errors in the following turns. Therefore, it is nontrivial to utilize the auxiliary generator, though not well trained when the seed data is limited, to complement belief states. With increasing seed data to train the auxiliary model, we can better detect belief state slots forgotten by GPT-3 and mitigate the \textit{de-generation} issue.
% reminding it to generate annotation in the following turns. 
% It can prevent the errors from propagating to following turns and leading to larger deviations. 
% It would be better if the auxiliary generator could generate correct slots not detected by GPT-3. 

\noindent\textbf{Slot-value Match Filter.~} $\hat{B}$ and $\bar{B}$ contain the belief states detected by GPT-3 and auxiliary generator and are complementary. We thus combine them. When the predictions of GPT-3 and auxiliary generators have overlapped slots, we take the slot value detected by GPT-3, i.e., GPT-3 has a higher priority.
We then filter out the \textit{over-generated} slots whose values couldn't be matched in the user utterance, resulting in the final belief state $B$. 

The auxiliary generator and slot-value match filter are used jointly to automatically detect and correct annotation errors, mitigating the \textit{de-generation} and \textit{over-generation} issues. Taking the example in Figure~\ref{fig:process} for illustration, the auxiliary model detects the correct belief state (\textit{hotel, area, don't care}), which is missing in $\hat{B}$. On the contrary, the slot value \textit{cheap} of the slot \textit{hotel-pricerange} couldn't be detected in the user utterance, and thus is removed from the belief state. The resulting final belief state $B$ is used to automatically retrieve the DB entry $Q$ from a pre-defined database. 

As for the system side, GPT-3 can generate the dialog act $\hat{A}$ by concatenating the generated user utterance $U$ and belief state $B$ with the prompt. 
% We also utilize the auxiliary TOD model $\mathcal{M}$ to generate its prediction conditioned on the dialogue context $X$ and the user utterance $U$, revised belief states $B$, and DB query result $Q$:
% \begin{equation}\label{eq:verifier}
%     \bar{A} = \mathcal{M}(X, U, B, Q).
% \end{equation}
We also utilize the auxiliary TOD model $\mathcal{M}$ to generate its prediction $\bar{A}$ conditioned on the dialogue context $X$ and the user utterance $U$, revised belief states $B$, and DB query result $Q$. To ensure some dialogue logic is followed and the database queried result is taken into account, we write some rules to filter out invalid dialog acts and decide the final dialog act annotation $A$, which is then concatenated with the prompt to continue generating the system response $S$. 
% We experimentally found that dialog acts generated by the auxiliary model is satisfying in most cases. 

In most cases, GPT-3's generation is acceptable without requiring revision, and we cannot guarantee that all the errors can be detected and corrected (we list the frequency in Appendix~\ref{sect:revision}). However, the automatic revision on the fly is still essential, as GPT-3 tends to imitate the errors in following turns. Therefore, each revision can not only correct the current error but also avoid numerous potential mistakes in the following turns. 
In addition, when there is enough data $\mathcal{D}_s$ to train the auxiliary TOD model $\mathcal{M}$, \textsc{Dialogic} plays a more significant role as user simulator to interact with the well-trained system $\mathcal{M}$.

\subsection{Turn-level Generation for DST}
For the dialogue state tracking (DST) task, the belief states at each step are an accumulation of previous steps. Any errors from earlier steps will propagate to later steps. In addition, when focusing on the DST task and belief state annotations, it is not necessary to generate them along with the whole dialogue and other annotations. To avoid error accumulation and unnecessary cost, we propose a method that only generates user utterances and corresponding belief states at a turn level. 

As shown in Figure~\ref{fig:dst}, for each turn of the seed dialogue, we will simulate a new user turn with turn-level belief states and user utterances as an alternative to the original turn to form a new dialogue. To preserve the consistency of dialogue flow, we generate turn-level belief states according to the dialog act of the previous turn:
\begin{itemize}
    \item \textit{Request} means that the system is requesting user's requirements on some attributes (slots), and the user is expected to answer the question by mentioning values for requested slots. We thus generated new turn-level belief states by selecting some or all of the requested slots and adding some other unmentioned slots.
    \item \textit{Reqmore} means that the system asks the user whether he wants service for other domains. Under this circumstance, we select an unmentioned domain and randomly select several slots in this domain to form a new turn-level belief state.
\end{itemize}
For the other dialog acts, we can randomly select unmentioned slots in the current domain. Some examples of these three kinds of dialog acts are provided in Appendix as Table~\ref{tab:dst_example}.
Given the new belief state, we prompt GPT-3 to generate the corresponding user utterance, which is then verified and revised as described in Section~\ref{sect:generation}. As for in-context examples, we only need to sample user turns instead of whole dialogues, which largely reduces the length of prompts and, thus, the generation cost.

Similarly, \citet{li2020coco} proposed a method that can only substitute the slot values of turn-level belief states to form new belief states (\textit{Value Substitution}), and trains GPT-2 to generate the corresponding user utterance. However, it requires a large set of data to train the utterance generator, which is not available in low-resource settings.

% \begin{table}[!t]
%     \small
% 	\centering 
% 	\scalebox{1}{
% 	\begin{tabular}{lccc}
% 		\toprule
% 		Method & Crowdsourcing* & \textsc{Dialogic} & Save\\
% 		\hline
% 		cost per dialog &   & \$0.52\\
% 	    cost per turn &  &\$0.06\\
%         \hline
% 		time per dialog &  & 25.2s\\
% 		time per turn &  & 3.1s\\
% 		\hline
% 		crowdworker & 1249 & 0 \\
% 		\bottomrule
% 	\end{tabular}}
%     \caption{Cost of creating MultiWOZ dataset via crowdsourcing~\citep{multiwoz} and our method \textsc{Dialogic}.}
% 	\label{tab:cost}
% 	\vspace{-4mm}
% \end{table}

\begin{table*}[!htbp]
    \small
	\centering 
	\renewcommand{\arraystretch}{1.2}
	\setlength{\tabcolsep}{6pt}
	\scalebox{0.86}{
	\begin{tabular}{cc|cccc|cccc|cccc|c}
	   % \hline
% 		\hlinewd{0.75pt}
		\toprule
		\multirow{2}{*}{\tabincell{c}{\textbf{Seed}\\\textbf{data}}}&\multirow{2}{*}{\tabincell{c}{\textbf{Augmented}\\\textbf{data}}}&\multicolumn{4}{c}{\textbf{SimpleTOD}}&\multicolumn{4}{c}{\textbf{MinTL-T5}}&\multicolumn{4}{c}{\textbf{PPTOD}}&\multirow{2}{*}{\tabincell{c}{\textbf{Avg.}\\\textbf{Improv.}}}\\
		\cmidrule(lr){3-6}
		\cmidrule(lr){7-10}
		\cmidrule(lr){11-14}
		&&I&S&B&C&I&S&B&C&I&S&B&C\\
		\hline
		\multirow{4}{*}{\tabincell{c}{1\%\\(85)}}&Base&37.94&25.53&6.40&38.13&56.81&40.38&12.16&60.76&55.36&38.44&12.25&59.14&-\\
		\cmidrule(lr){2-15}
		&+Orig.(85)&46.45&33.73&\textbf{6.96}&47.05&64.93&50.20&12.37&70.13&60.26&\textbf{44.94}&\textbf{12.85}&65.45&16.50\%\\
		&+Sim.(85)&\textbf{49.65}&\textbf{36.74}&6.90&\textbf{50.09}&\textbf{69.44}&\textbf{50.30}&\textbf{12.46}&\textbf{72.33}&\textbf{68.67}&44.54&12.82&\textbf{69.43}&\textbf{22.60\%}\\
		\hline
		\multirow{4}{*}{\tabincell{c}{5\%\\(422)}}&Base&55.96&42.44&7.66&56.86&74.05&60.42&14.71&82.70&68.17&53.55&13.58&74.44&-\\
		\cmidrule(lr){2-15}
		&+Orig.(422)&57.96&46.85&\textbf{8.04}&60.44&72.24&60.42&\textbf{14.91}&81.24&72.07&\textbf{61.26}&\textbf{15.94}&\textbf{82.61}&5.17\%\\
		&+Sim.(422)&\textbf{58.96}&\textbf{47.35}&7.86&\textbf{61.02}&\textbf{77.45}&\textbf{64.93}&13.98&\textbf{85.17}&\textbf{74.67}&60.06&14.20&81.56&\textbf{6.62\%}\\
		\hline
		\multirow{4}{*}{\tabincell{c}{10\%\\(843)}}&Base&57.96&46.85&8.04&60.44&72.24&60.42&14.91&81.24&72.07&61.26&15.94&82.61&-\\
		\cmidrule(lr){2-15}
		&+Orig.(843)&59.26&47.45&\textbf{8.30}&61.65&78.76&68.74&\textbf{15.92}&89.67&\textbf{78.48}&\textbf{64.46}&\textbf{15.22}&\textbf{86.69}&5.77\%\\
		&+Sim.(843)&\textbf{62.86}&\textbf{51.75}&8.16&\textbf{65.47}&\textbf{79.96}&\textbf{69.84}&15.41&\textbf{90.31}&77.98&63.36&14.39&85.06&\textbf{7.48\%}\\
		\hline
		\multirow{4}{*}{\tabincell{c}{100\%\\(8438)}}&Base&68.67&61.05&10.21&75.08&80.06&72.85&17.87&94.33&\textbf{83.88}&69.47&16.33&93.01&-\\
		\cmidrule(lr){2-15}
		&+Sim.(422)&68.38&61.82&\textbf{10.32}&75.42&79.46&73.45&18.52&94.98&82.88&70.97&\textbf{19.22}&\textbf{96.15}&1.57\%\\
		&+Sim.(843)&\textbf{68.97}&\textbf{62.46}&10.21&\textbf{75.93}&\textbf{80.76}&\textbf{74.15}&\textbf{18.72}&\textbf{96.18}&83.08&\textbf{71.77}&18.44&95.87&\textbf{2.12\%}\\
        \bottomrule
	\end{tabular}}
    \caption{End-to-end dialogue modeling evaluation on MultiWOZ2.3~\citep{multiwoz23}, where I, S, B, C stand for the Inform, Success, BLEU, and Combined Score metrics, respectively. Sim. and Orig. stand for simulated and original dialogues. The highest scores are bolded. The average improvements are w.r.t. the combined scores.}
	\label{tab:experiment}
	\vspace{-4mm}
\end{table*}

\section{Experiments}
\subsection{Experimental Setup}
\noindent\textbf{Seed Dataset.~}
We implemented our method on the  MultiWOZ~\citep{multiwoz} dataset, which consists of 8,438 training, 1,000 validation, and 1,000 test dialogues across 7 domains. As annotation errors exist in the original dataset, we conduct experiments on a cleaner version MultiWOZ2.3~\citep{multiwoz23}. To simulate the challenging low-resource scenarios, we use 1\% (85/8438), 5\% (422/8438), and 10\% (843/8438) as the seed training dataset and adopt the standard val/test set for evaluation. We also simulate 422/843 dialogues given the full training set to evaluate its effectiveness as a data augmentation method under the full-shot setting.

\noindent\textbf{Simulated Dataset.~}
We select the largest version of GPT-3 API \textit{text-davinci-002} and use the top-$p$ decoding, where $p=0.7$. When generating the user goal, we limit the maximum number of requested domains in a dialogue to 4 and the maximum of slots in each domain to 6. We stop generating a dialogue if the number of turns exceeds 12. We use two in-context examples for all generations. More details are provided in Appendix~\ref{sect:dialog_generation}. PPTOD trained on the seed dataset is utilized as the auxiliary model for automatic revision. 
% The cost for dialogue generation and user turn generation is presented in Table~\ref{tab:cost}.

\noindent\textbf{Cost Comparison.~}
MultiWOZ dataset creation required 1,249 workers and cost around \$30k except for the additional cost of postprocessing~\citep{multiwoz}. Assuming a minimum hourly wage of \$8, the whole process would take up to 3,750 work hours. In comparison, our approach doesn't require either \textit{human involvement} or \textit{parameter update}, except for the minimal efforts in collecting the small seed dataset and training an auxiliary model on it. The cost and time are thus mainly derived from GPT-3 API call.\footnote{https://openai.com/api/pricing/} Averagely, generating a dialogue using GPT-3 only cost \textbf{\$0.52}, while generating a training sample (turn) for DST augmentation only cost \textbf{\$0.006}. Using other open-resourced LLMs such as OPT-175B~\citep{zhang2022opt} can avoid the cost and make our method almost free. Each dialogue can be generated within a few seconds, meaning we can create a large-scale dataset such as MultiWOZ within several hours, which largely shortens the time for dataset creation.

% Therefore, we can create a large-scale and high-quality dataset like MultiWOZ in several hours. 
% And the generation of a user turn for DST augmentation only cost \$0.06.
% Each dialogue turn costs \textbf{\$0.06} and \textbf{3.1} seconds. 
% By contrast, Given the total amount of 10,438 dialogues and assuming a minimum hourly wage of \$6 for the crowdworkers, generating each dialogue would averagely cost at least 0.48 work hour and \$2.87. Therefore, \textsc{Dialogic} reduces the cost by nearly \textbf{60 times} in time and \textbf{6 times} in money.

% Specifically, generating a dialogue in it averagely costs \textbf{\$0.52} and \textbf{\$0.06}, respectively. Generating a user turn for DST augmentation only costs \textbf{\$0.0003}. The generation of a dialogue can be finished in a few seconds. On the contrary, generating a dialogue via crowdsourcing is much more expensive and slower.

\noindent\textbf{Evaluation Metric.~}
To assess the quality of the simulated dialogues, we evaluate the performance of models trained on these simulated dialogues on two benchmark TOD tasks: (1) end-to-end dialogue modeling (E2E) and (2) dialogue state tracking (DST). For E2E evaluation, we use the metrics defined in MultiWOZ~\citep{multiwoz}: \textit{Inform}, \textit{Success}, \textit{BLEU}, and an overall metric \textit{Combined Score}: $BLEU+0.5\times(Inform+Success)$. For DST evaluation, we report the joint accuracy.

\noindent\textbf{Baselines.~}
We select the following three recent end-to-end TOD models as baseline models: SimpleTOD~\citep{simpletod}, MinTL~\citep{mintl}, and PPTOD~\citep{pptod}. These three models are all based on pre-trained transformers. SimpleTOD is initialized with $\text{GPT-2}_{\text{small}}$, while MinTL and PPTOD are initialized with $\text{T5}_{\text{small}}$. PPTOD has also been pretrained on heterogeneous dialogue corpus, making it more powerful in low-resource settings. These three models are all capable of performing end-to-end dialogue modeling tasks and DST tasks. We also experiment on a classic DST model TRADE~\citep{trade}. For a fair comparison, we use the delexicalized system response in the same format and the evaluation script as in~\citep{damd,mintl,pptod} for E2E evaluation on all these models. During inference, we didn't use any oracle information. 
% The DST prediction is used to query the database, while the generated dialog action and system response are used as dialogue context. 
For DST evaluation, we use lexicalized utterances.
We use the default hyperparameters in their original implementations.

\subsection{Experimental Results}

\begin{table}[!t]
    \small
	\centering 
	\renewcommand{\arraystretch}{1.0}
	\setlength{\tabcolsep}{5pt}
	\scalebox{0.88}{
	\begin{tabular}{lcccccc}
	   % \hline
% 		\hlinewd{0.75pt}
        \toprule
		\multirow{2}{*}{Statistics}&\multicolumn{2}{c}{1\%}&\multicolumn{2}{c}{5\%}&\multicolumn{2}{c}{10\%}\\
		\cmidrule(lr){2-3}
		\cmidrule(lr){4-5}
		\cmidrule(lr){6-7}
		&Sim.&Orig.&Sim.&Orig.&Sim.&Orig.\\
		\hline
		Total dialogs &85&85&422&422&843&843 \\
		Total turns &616&599&3510&2778&6634&5617\\
% 		Total tokens &&-&-&-&-&-\\
		Total domains &229&147&1076&738&2203&1471\\
% 		Tot. offered entities &90&65&370&287&810&588\\
% % 		Offer rate (\%) &38.96\%&44.22\%&34.32\%&38.89\%&37.04\%&39.97\%\\
% 	    Offer rate (\%) &39&44&34&39&37&40\\
		Avg. turns &7.25&7.05&8.32&6.58&7.87&6.66\\
% 		Avg. tokens &-&-&-&-&-&-\\
		Avg. domains &2.69&1.73&2.55&1.75&2.61&1.74\\
		Uniq. tokens &550&561&1095&1084&1336&1373\\
		Uniq. 3-grams &3413&4300&10662&13815&15063&22261\\
% 		Goal diversity &-&-&-&-&-&-\\
% 		\hlinewd{0.75pt}
        \bottomrule
    	\end{tabular}}
    \caption{Comparison of the statistics of simulated dialogues (Sim.) with human generated dialogues from the original MultiWOZ dataset (Orig.).}
	\label{tab:diversity}
	\vspace{-5mm}
\end{table}

\subsubsection{End-to-end Dialogue Modeling}\label{sect:results}
We here investigate a realistic question when building a TOD system for a new task or domain: \textit{would our method be a good alternative to crowdsourcing in expanding a small corpus of dialogue data?} To answer this question, we combine the seed dataset with (1) simulated dialogues with our method from the seed dataset (Sim.); (2) human-generated dialogues from the original dataset, excluding the seed data (Orig.). We then train several representative TOD models with these two datasets and compare their performance. For a fair comparison, we use the \textbf{same amount} of simulated dialogues and original dialogues and the same set of seed data across all setups. 

As seen in Table~\ref{tab:experiment}, the models trained on simulated dialogues along with the seed dataset perform much better than those only trained on the seed dataset. 
% The performance improvement is especially significant with in the most challenging 1\% setting, which can be attributed to the powerful few-shot learning ability of GPT-3. 
Remarkably, compared with human-generated dialogues, the same amount of our simulated dialogues can lead to even more significant performance improvement in most cases. 
% In a detailed manner, the Inform and Success rate of models trained on simulated dialogues are consistently better than those on human-generated dialogues. An opposite trend is seen in BLEU scores. 
Our simulated dialogues can still improve performance when the full training data are provided. We only show the result of using a small amount of simulated dialogues (422/843) here. One can also generate more dialogues to further improve the performance. The results suggest the effectiveness of our method as a data augmentation method. This is expected as when we have more seed data to select from as in-context examples for GPT-3 and train the auxiliary revision model, we can generate more diverse dialogues, more accurate annotations, and thus dialogues with better quality.
% As a result, our method can not only be used as a 
% , which indicates that our method can generate more varied dialogue flows but less diverse utterances.

To understand the observations, we further analyze the statistics of simulated and human-generated dialogues and present them in Table~\ref{tab:diversity}. We found that our simulated dialogues have more requested domains and dialogue turns compared with human-generated dialogues, which are controlled by the generated user goals. The much more sub-tasks and sub-conversations in the simulated dialogues improve the model's ability to deal with more complex and challenging multi-domain tasks, thus leading to a higher Inform rate and Success rate. On the contrary, the system responses generated by our method have a comparable amount of unique tokens but fewer 3-grams than the original ones, which explains why the BLEU score is slightly lower than the human-generated dialogues.

\begin{table}[!t]
    \small
	\centering 
	\scalebox{1}{
	\begin{tabular}{lcccccccccccc}
	   % \hline
% 		\hlinewd{0.75pt}
        \toprule
% 		Model &\multicolumn{2}{c}{Setup}&\multicolumn{4}{c}{1\%}&\multicolumn{4}{c}{5\%}&\multicolumn{4}{c}{10\%}\\
% 		\cmidrule(lr){2-3}
% 		\cmidrule(lr){4-7}
% 		\cmidrule(lr){8-11}
% 		\cmidrule(lr){12-15}
		Model&I&S&B&C\\
		\hline
		SimpleTOD&49.65&36.74&6.90&50.09\\
		\quad\textit{w/o revision}&43.92&29.13&6.21&42.74\\
		\hline
		MinTL-T5&69.44&50.30&12.46&72.33\\
	    \quad\textit{w/o revision}&61.68&46.34&11.83&65.84\\
		\hline
		PPTOD&68.67&44.54&12.82&69.43\\
	\quad\textit{w/o revision}&62.41&40.27&11.97&63.31\\
    \bottomrule
	\end{tabular}}
    \caption{Ablation study on the automatic revision under 1\% low-resource setting.}
	\label{tab:ablation}
	\vspace{-4mm}
\end{table}

\noindent\textbf{Effect of Automatic Revision.~}
Table~\ref{tab:ablation} shows performances of PPTOD trained with dialogues generated in the presence and absence of the automatic revision under the 1\% setting. Without automatic revision, the generated dialogues lead to lower performance improvement in all metrics, which suggests the importance of our controllable generation strategy and automatic revision methods. We list the frequency of the automatic revision in Appendix~\ref{sect:revision}.
% This is reasonable as GPT-3 sometimes generates incorrect belief states and dialog acts, declining the model performance. What's worse, GPT-3 will imitate these errors in consecutive turns if not corrected, leading to more significant problems. 
% This suggests the importance of the controllable generation and automatic revision methods.

\subsubsection{Dialogue State Tracking}
Next, we investigate the effectiveness of belief state annotations generated by our method as augmented data for DST training in low resources settings. We simulate the low-resource setting using 1\% of the MultiWOZ training set as seed data. Table~\ref{tab:dst_experiment} shows the performance of various models trained on different sizes of augmented data. The augmented data can consistently improve the model performance across different models. With the increase of augmented data, the accuracy keeps increasing, though the upward trend is gradually slowing down.
% Considering the low generation cost, we can draw the conclusion that \textsc{SimDial}-DST is an efficient and effective data augmentation method for DST.

\begin{table}[!t]
    \small
	\centering 
	\renewcommand{\arraystretch}{1.0}
	\setlength{\tabcolsep}{5pt}
	\scalebox{1}{
	\begin{tabular}{lccccc}
	   % \hline
% 		\hlinewd{0.75pt}
        \toprule
		\multirow{2}{*}{Model}&\multirow{2}{*}{Base}&\multicolumn{4}{c}{Augmentation size}\\
		\cmidrule(lr){3-6}
		&&x1&x2&x3&x4\\
		\hline
		TRADE &12.98&14.41&15.96&16.60&17.17 \\
		SimpleTOD &13.69&16.43&17.81&18.51&19.15\\
% 		Total tokens &&-&-&-&-&-\\
		MinTL &23.25&27.30&27.69&28.71&29.27\\
	    PPTOD &34.48&36.71&37.74&38.12&38.51\\
% 		\hlinewd{0.75pt}
        \bottomrule
    	\end{tabular}}
    \caption{DST evaluation with 1\% of training data and different sizes of augmented data.}
	\label{tab:dst_experiment}
	\vspace{-4mm}
\end{table}

\subsection{Human Evaluation}
To get a more comprehensive measure of the quality of the simulated data compared with the original human-generated data, we conducted a blind human evaluation study. Three participants with NLP backgrounds are given 50 dialogues simulated from 1\% of the training data and another 50 dialogues from the original dataset without knowing their source (simulated or original). Following~\citep{Mohapatra2020SimulatedCF}, we ask the participants to check the quality of the conversation and annotations of each dialogue turn by answering the following questions: (1) ``\textit{Are the utterances grammatically correct?}'' (2) ``\textit{Is the user utterance fluent and natural?}'' (3) ``\textit{Is the system response fluent and natural?}'' (4) ``\textit{Is the belief state annotation consistent with the user utterance?}'' (5) ``\textit{Is the dialog act consistent with the system response?}''. For each question, the participants should answer ``yes'' or ``no''. We adopt a majority vote approach to decide the final answer with at least two votes. Table \ref{tab:human_evaluation} shows the percentage of dialogue turns that satisfy each quality measure (with the answer of ``yes'').

We find that our method can generate even more grammatically correct conversations than human crowdworkers leveraging the strong generation ability of GPT-3. The generated user utterances are comparably fluent and natural to original human-generated ones. In contrast, the generated system responses are not that fluent. We suspect this is because the system responses are delexicalized, which is more challenging for GPT-3 to understand and imitate. As for the annotation quality, although humans generate original dialogues, annotation errors still exist, suggesting the difficulty of task-oriented dialogue annotations. As expected, our generated dialogue data has more annotation errors than human-generated ones. Considering only as few as 85 dialogues are given and no human involvement is required, we believe the gap is acceptable and can be bridged with more seed data provided.
% As we use an auxiliary model trained on limited data to correct potential errors automatically instead of costly human revision. 
% There is a trade-off between the annotation quality and required human efforts. 
% The gap between simulated dialogues and the original dialogues on the user consistency is not as large as that on the system consistency because the belief states are easier for GPT-3 to understand and generate compared with the more complex dialog acts. 
Overall, the conversations generated by our method have comparable quality to human-generated ones. However, the generated annotations are not as accurate as human annotations. Satisfyingly, the noisy annotation still introduces considerable performance improvement.

\begin{table}[!t]
    \small
	\centering 
	\renewcommand{\arraystretch}{1.2}
	\setlength{\tabcolsep}{6pt}
	\scalebox{1}{
	\begin{tabular}{lcc}
	   % \hline
% 		\hlinewd{0.75pt}
        \toprule
		Measure&Simulated&Original\\
		\hline
		Grammar&99.00&96.32\\
		User fluency&94.55&95.22\\
		System fluency&88.61&94.48\\
		\hline
		User consistency&84.65&96.69\\
		System consistency&73.76&94.12\\
% 		\hlinewd{0.75pt}
        \bottomrule
	\end{tabular}}
    \caption{Human evaluation on the simulated dialogues and original dialogues w.r.t. the percentage of dialogue turns that satisfy each quality measure.}
	\label{tab:human_evaluation}
	\vspace{-4mm}
\end{table}

\section{Conclusion}
In this paper, we propose a dialogue simulation method based on large language model in-context learning to automate dataset creation. Our proposed method can generate dialogues and annotations given only a few seed dialogues. The simulation process requires zero or minimum \textit{human involvement} and \textit{model training}, making our method much more cost-efficient and time-saving than crowdsourcing. Human and automatic evaluations demonstrate that the simulated dialogues have comparable quality to human-generated ones, which shows the potential of our method as an alternative to crowdsourcing in dialogue dataset creation.

\section*{Limitations}
In this paper, we investigate ways to leverage in-context learning with GPT-3 to automatically generate high-quality task-oriented dialogues for building dialogue systems.
Although our method can already generate high-quality dialogues without requiring human involvement, there are still some limitations in real-world applications. GPT-3 has a deficiency in lack of reliability and is inevitable to generate some unexpected data even with automatic revision. Human review and revision are still necessary to ensure the annotation is completely correct. However, it is challenging as the revision at each step will influence the latter steps. Therefore, an effective and efficient human and machine collaboration approach is our future direction. In addition, as the dialogues along with annotations are very lengthy, it is essential to reduce their length to lower the generation cost and enable the use of more in-context examples.

\section*{Ethics Statement}
Our proposed method instructs LLMs to generate dialogues for building dialogue systems. However, LLMs such as GPT-3~\citep{gpt3} are observed to generate toxic or biased text \citep{gpt3,lucy-bamman-2021-gender,chan2022gpt}. 
Although a new version of GPT-3 called InstructGPT has been released, trying to reduce these toxic languages, the issue hasn't been sufficiently addressed.
Thus, automatic filtering or human review methods is necessary to exclude some parts of training data to avoid the models generating undesirable responses containing toxicity and bias from the simulated dialogues.

\section*{Acknowledgements}
The authors would like to thank the anonymous reviewers for their thoughtful comments. This research was partly sponsored by the DARPA PTG program (HR001122C0009) and Alexa Prize Taskbot Challenge. Any opinions, findings, conclusions, or recommendations expressed in this paper are those of the authors and do not necessarily reflect the views of funding agencies.

\bibliography{anthology,custom}
\bibliographystyle{acl_natbib}

\newpage
\appendix
\section*{Appendix}

\begin{table*}[!ht]
\centering
\caption{Full ontology for all domains in MultiWOZ2.3~\cite{multiwoz23}. The upper script indicates which domains it belongs to. *: universal, 1: restaurant, 2: hotel, 3: attraction, 4: taxi, 5: train, 6: hospital, 7: police. The table is adapted from \cite{multiwoz} with some modification.}
\begin{tabular}{|l|l|}
\hline
dialog acts & \begin{tabular}[c]{@{}l@{}}$\text{inform}^{*}$ / $\text{request}^{*}$ / $\text{select}^{1235}$ / $\text{recommend/}^{123}$ / $\text{nooffer}^{1235}$ / $\text{offerbook}^{125}$ / \\ $\text{offerbooked}^{125}$ / $\text{nobook}^{12}$ /
$\text{welcome}^{*}$ / $\text{greet}^{*}$ / $\text{bye}^{*}$  / $\text{reqmore}^{*}$  \end{tabular}                                                                                                                                      \\ \hline
slots    & \begin{tabular}[c]{@{}l@{}}$\text{address}^{12367}$ / $\text{postcode}^{1367}$ / $\text{phone}^{123467}$ / $\text{name}^{123}$ / $\text{area}^{123}$ / $\text{pricerange}^{12}$ / \\ $\text{type}^{23}$ /  $\text{internet}^{2}$ / $\text{parking}^{2}$ / $\text{stars}^{2}$ / $\text{departure}^{45}$ /  $\text{destination}^{45}$ / $\text{leave}^{45}$ / \\ 
$\text{arrive}^{45}$ / $\text{people}^{125}$ / $\text{reference}^{1235}$ / $\text{id}^{5}$ / $\text{price}^{5}$ / $\text{time}^{15}$ / $\text{department}^{6}$ / \\
$\text{day}^{125}$ / $\text{stay}^{2}$ / $\text{car}^{4}$ / $\text{food}^{1}$ 
\end{tabular} \\ \hline
\end{tabular}
%\vspace*{-1em}
  \label{tab:ontology}
\end{table*}

\section{Implementation Details}\label{sect:implementation}
We conduct experiments on MultiWOZ2.3\footnote{https://github.com/lexmen318/MultiWOZ-coref}, which corrects annotations in dialogue acts from dialogue states and adds co-reference annotation. The ontology information of MultiWOZ2.3 dataset is presented in Table~\ref{tab:ontology}. There are 12 dialog acts and 24 slots in total. We use the released source codes and default hyperparameters of the baseline models SimpleTOD\footnote{https://github.com/salesforce/simpletod}, MinTL\footnote{https://github.com/zlinao/MinTL}, PPTOD\footnote{https://github.com/awslabs/pptod}, and TRADE\footnote{https://github.com/jasonwu0731/trade-dst}. For a fair comparison, we use the same set of training data and dev/test data across all these models. In addition, we use the same evaluation script provided in PPTOD for all these models. The delexicalized utterances are used for end-to-end evaluations, while lexicalized utterances are used for DST evaluation. All the experiments are run on a server with 8 NVIDIA RTX A6000 GPUs.

% \begin{table}[!t]
% \small
% \centering
% \scalebox{1.0}{
%     \begin{tabular}{l c c}
%     \toprule[1.2pt]
%     \textbf{Hyper-parameter} & \textbf{Dialogue-level} & \textbf{Turn-level}\\
%     \hline
%     temperature & $0.7$ & $0.7$ \\ 
%     top\_p & $1.0$ & $1.0$ \\
%     max\_output\_tokens & $64$ & $64$ \\
%     frequent\_penalty & $1.0$ & $1.0$ \\
%     presence_penalty & 0.0 & 0.0 \\
%     num\_shots & $2$ & $2$ \\
%     \bottomrule[1.2pt]
%     \end{tabular} 
% }
% \caption{Hyper-parameters we set for the dialogue generation and turn-level generation for DST augmentation.}
% \label{tab:hyper-parameters}
% \end{table}

\subsection{Dialogue Generation}\label{sect:dialog_generation}
We select the most powerful version of GPT-3 API \textit{text-davinci-002}\footnote{https://openai.com/api/pricing/}. 
% The hyper-parameters we use are presented in Table~\ref{tab:hyper-parameters}. 
The temperature is set as 0.7, the top\_p and frequency penalty are both set as 1.0 to encourage the generation of more diverse data and avoid the repetition of the in-context examples. We use $2$ shots for demonstrations. When generating the dialogues, we set the maximum number of turns as 12 to avoid endless generations. 

We test three strategies to generate the user goals and prompt: (1) \textit{Random Sampling}; (2) \textit{Value Substitution}; and (3) \textit{Combination}. For \textit{Random Sampling} strategy, the sampling distribution of the number of domains, the minimal number of slots in each domain, and the maximum number of slots in each domain are presented in Table~\ref{tab:rs}. For the \textit{Random Sampling} strategy, we keep all the slots and substitute the slots values of the original user goal to create new goals. For the \textit{Combination} strategy, we set the temperature in Equation~\ref{eq:sampling} as 0.2 when sampling similar dialogues from the seed dataset.

\begin{table}[!h]
    \small
	\centering 
	\scalebox{1}{
	\begin{tabular}{ccc|c}
	    \hline
		\#domains & Min \#slots &Max \#slots &Probability\\
		\hline
		1& 4& 6&0.3\\
		2& 3& 5&0.6\\
		3& 2& 5&0.1\\
		\hline
	\end{tabular}}
    \caption{The sampling distribution of user goals w.r.t. the number of domains, the minimal and maximum number of slots in each domain for the  \textit{Random Sampling} strategy.}
	\label{tab:rs}
% 	\vspace{-2mm}
\end{table}

\begin{figure}[h!]
    \centering
    \includegraphics[width=0.9\linewidth]{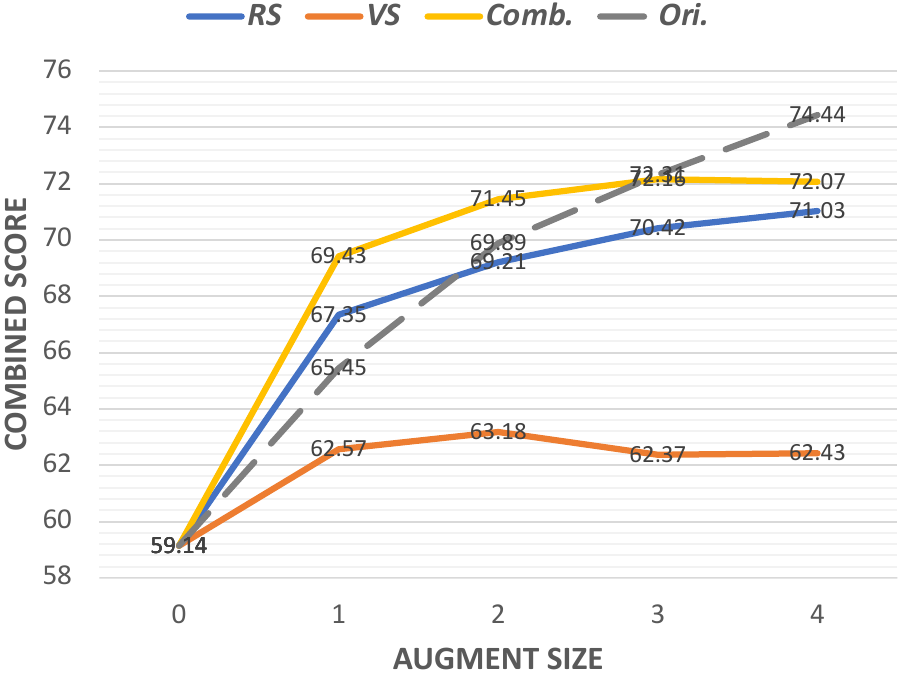}
    \caption{The combined score of PPTOD trained with dialogues simulated with different methods and augmentation size with 1\% training data as the seed data.}
    \label{fig:augment_size}
    \vspace{-2mm}
\end{figure}

\subsection{Goal Generation Methods}\label{sect:augmentation_method}
As mentioned in Section~\ref{sect:example}, we investigate the following methods to generate user goals and retrieve dialogues as in-context examples: (1) \textit{Random Sampling (RS)}, (2) \textit{Value Substitution (VS)}, and (3) \textit{Combination (Comb.)}. 
We show the performance of PPTOD trained on dialogues simulated with different goal generation methods and varying augmentation sizes in Figure~\ref{fig:augment_size}. We also show the model performance trained on the same amount of original dialogues in the dashed line as a baseline (Ori.). 

We can see that the \textit{VS} method performs worst, as it can only change the slot values. Thus, GPT-3 tends to simply replicate the in-context examples. As the \textit{Combination} method can cover most of the information needed in the target dialogues and encourage GPT-3 to generate diverse data, it achieves the best performance. However, it is harder to improve further with the increased augmentation size as there are limited seed dialogues and, thus, their combinations. On the contrary, as the \textit{RS} method can generate any user goals, the model performance keeps increasing. However, there is still an upper bound of performance, which depends on the number of provided seed dialogues.

\subsection{Turn-level Generation}~\label{sect:dst_generation} We generate user turns for DST augmentation based on the human-generated seed dialogues. The hyper-parameters for GPT-3 API call are the same as dialogue generation as in Appendix~\ref{sect:dialog_generation}. We use $2$ in-context examples by default. One can also increase the number for better generation quality. For each turn from the original dialogues, we generate augmented user turns and concatenate them with the previous turns in this dialogue to create a new training sample. We generate the user turns based on the dialog acts of the last system turn in the context. Suppose the last system turn contains the \textit{request} act. In that case, we will randomly select at least one slot from the requestable slots and another at least two slots from the unmentioned slots in the current domain to form the belief state of the augmented user turn. If the last system turn contains the \textit{reqmore} act, we will randomly select one unmentioned domain and at least $1$, at most $4$ slots in the domain. For the other cases, we will randomly drop at least one slot from the original belief state and add at least one unmentioned slot. Having selected the slots in the belief state, we will randomly select a possible slot value for each slot to create the concrete belief state for the augmented user turn.

\section{Automatic Revision Frequency}\label{sect:revision}
To investigate how often the data generated by GPT-3 need to be revised and how many errors can be corrected by our method, we randomly selected 20 dialogues simulated under the 1\% low-resource setting and manually checked the annotations. In a total of 170 turns, GPT-3 generates incorrect belief state annotations in 31 turns (18 \textit{de-generations} / 13 \textit{over-generations}). The auxiliary generator corrects GPT-3 in 13 out of 18 de-generated turns, while the slot-value match filter corrects in 10 out of 13 over-generated turns. Finally, only 6.47\% of the revised belief states are still incorrect (11 out of 170), while that for dialog acts is 11.18\% (19 out of 170).

\section{Generation Examples}

\subsection{Dialogue Generation Example}\label{sect:dialog_example}
An example of a complete prompt is shown in Table~\ref{tab:whole_prompt}, given which the controllable generation process of the dialogue is presented in Table~\ref{tab:whole_generation_process}. The process is fully automated.

\begin{table*}[htb!]
    \centering
    \scriptsize
    \renewcommand\arraystretch{1.6}
        % \resizebox{\textwidth}{70mm}{
    \begin{tabular}{c|p{13.5cm}}
    % \hline
    \specialrule{.3em}{.2em}{.2em}
        \tabincell{c}{\textbf{Task}\\\textbf{Description}} & \tabincell{l}{The following are conversations between a user and an assistant.  The assistant can help the user to find things that satisfy his requirements.\\ Try to speak differently in different conversations.}
 \\
        \hline
        \tabincell{c}{\textbf{Example 1}\\\textbf{(PMUL1576)}} & \tabincell{l}{\textbf{Instruction1:} You are going to book a train, and your requirements for the train are (\textcolor{red}{[train] destination is leicester , departure is} \\ \textcolor{red}{cambridge , leave is 08:45 , day is saturday , arrive is dontcare}). You also want to book a hotel, and your requirements for the hotel \\ are (\textcolor{red}{[hotel] name is cityroomz , stay is 4 , day is tuesday , people is 8}). Make sure you get the booking information once booked.\\
        \textbf{Conversation1:} \\
        \textbf{User}(\textcolor{red}{[train] destination is leicester , departure is cambridge , leave is 08:45}): \textcolor{darkblue}{i really need to get out of cambridge ! can you find } \\ \textcolor{darkblue}{me a train to leicester ? sometime after 08:45 because i like to sleep in .} \\
        \textbf{Assistant}(\textcolor{red}{[train] [request] day}): \textcolor{darkgreen}{i would be happy to help with your request , what day will you be leaving ?} \\
        \textbf{User}(\textcolor{red}{[train] day is saturday}): \textcolor{darkblue}{i 'll be leaving this place on saturday .} \\
        \textbf{Assistant}(\textcolor{red}{[train] [request] arrive}): \textcolor{darkgreen}{what time would you like to arrive by ?}\\
        \textbf{User}(\textcolor{red}{[train] arrive is dontcare}): \textcolor{darkblue}{it does not matter .}\\
        \textbf{Assistant}(\textcolor{red}{[train] [inform] destination arrive leave [offerbook]}): \textcolor{darkgreen}{ there is a train that leaves at [value_leave] and arrive in leiester} \\ \textcolor{darkgreen}{at [value_arrive] . would you like me to book it for you ?} \\
        \textbf{User}(\textcolor{red}{[train]}): \textcolor{darkblue}{no thank you . what is the cost of the ticket ?} \\
        \textbf{Assistant}(\textcolor{red}{[train] [inform] price [general] [reqmore]}): \textcolor{darkgreen}{the ticket price is [value_price] . can i be of further assistance ?} \\
        \textbf{User}(\textcolor{red}{[train]}): \textcolor{darkblue}{i also need the travel time and arrival time please .}\\
        \textbf{Assistant}(\textcolor{red}{[train] [inform] arrive time}): \textcolor{darkgreen}{arrival time is [value_arrive] and travel time is [value_time] . }\\
        \textbf{User}(\textcolor{red}{[hotel] name is cityroomz}): \textcolor{darkblue}{i ' m also looking for a particular hotel . its name is called cityroomz .}\\
        \textbf{Assistant}(): \textcolor{darkgreen}{sure , what kind of information do you need ?}\\
        \textbf{User}(\textcolor{red}{[hotel] stay is 4 , day is tuesday , people is 8}): \textcolor{darkblue}{i would like to to book it for 8 people and 4 nights starting from tuesday}\\
        \textbf{Assistant}(\textcolor{red}{[hotel] [offerbooked] day people reference stay [general] [reqmore]}): \textcolor{darkgreen}{i have booked a room for [value_people] for [value_stay] }\\ \textcolor{darkgreen}{nights beginning on [value_day] . your reference number is [value_reference] . is there anything else i can help you with today ?}\\
        \textbf{User}(\textcolor{red}{[general]}): \textcolor{darkblue}{not at this time . thank you .}\\
        \textbf{Assistant}(\textcolor{red}{[general] [bye]}): \textcolor{darkgreen}{have a fantastic day , goodbye .}}
 \\
        \hline
        \tabincell{c}{\textbf{Example 2}\\\textbf{(SNG0955)}} & \tabincell{l}{\textbf{Instruction2:} You are going to book a hotel, and your requirements for the hotel are (\textcolor{red}{hotel] pricerange is expensive , area is east ,}\\ \textcolor{red}{parking is yes}). Make sure you get the booking information once booked.\\
        \textbf{Conversation2:} \\
        \textbf{User}(\textcolor{red}{([hotel] pricerange is expensive}): \textcolor{darkblue}{i need a place to stay that does n't have to have internet and is in the expensive price range please .} \\ 
        \textbf{Assistant}(\textcolor{red}{([hotel] [inform] choice [request] area}): \textcolor{darkgreen}{i have [value_choice] different ones all around town . did you prefer to stay in a certain}\\\textcolor{darkgreen}{area ?} \\
        \textbf{User}(\textcolor{red}{([hotel] area is east}): \textcolor{darkblue}{yes , on the east side please .} \\
        \textbf{Assistant}(\textcolor{red}{[hotel] [inform] name [offerbook]}): \textcolor{darkgreen}{[value_name] meets your needs , would you like to book it ?}\\
        \textbf{User}(\textcolor{red}{[hotel] parking is yes}): \textcolor{darkblue}{does it have free parking ?}\\
        \textbf{Assistant}(\textcolor{red}{[hotel] [inform] parking}): \textcolor{darkgreen}{yes , it does .} \\
        \textbf{User}(\textcolor{red}{[hotel]}): \textcolor{darkblue}{i ' m not ready to book . can you just tell me what the address is ? oh , and how many stars is it ?} \\
        \textbf{Assistant}(\textcolor{red}{[hotel] [inform] type stars address [general] [reqmore]}): \textcolor{darkgreen}{sure . it is a [value_stars] star [value_type] and the address is }\\ \textcolor{darkgreen}{[value_address] . anything else ?} \\
        \textbf{User}(\textcolor{red}{[general]}): \textcolor{darkblue}{no , that is all . thanks .}\\
        \textbf{Assistant}(\textcolor{red}{[general][bye]}): \textcolor{darkgreen}{you are welcome ! please contact us if you would like to make a reservation in the future .}}
\\
        \hline
        \tabincell{c}{\textbf{Target}\\\textbf{Dialogue}\\\textbf{Prompt}} & \tabincell{l}{\textbf{Instruction3:} You are going to book a hotel, and your requirements for the hotel are (\textcolor{red}{([hotel] area is south , stay is 5 , people is 4}).\\ You also want to book a train, and your requirements for the train are (\textcolor{red}{[train] destination is birmingham new street , arrive is 13:06}). \\Make sure you get the booking information once booked.\\
        \textbf{Conversation3:} \\}
\\
    % \hline
    \specialrule{.3em}{.2em}{.2em}
    \end{tabular}
    % }
    \vspace{1.5mm}
    \caption{Example of a complete prompt which contains a task description, two in-context examples (PMUL1576 and SNG0955 are from the MultiWOZ dataset), and the user goal (\textit{instruction}) of the target dialogue. The \textcolor{red}{user goal}, \textcolor{red}{belief states}, and \textcolor{red}{dialog acts} are in red color. The \textcolor{darkblue}{user utterances} are in blue, while the \textcolor{darkgreen}{system responses} are in green. Given the prompt, we can generate the \textit{conversation} of the target dialogue as shown in Table~\ref{tab:whole_generation_process}. }
    \label{tab:whole_prompt}
\end{table*}

\begin{table*}[htb!]
\vspace{-18mm}
    \centering
    \scriptsize
    \renewcommand\arraystretch{1.5}
    % \resizebox{\textwidth}{90mm}{
    % \scalebox{1.2}{
    \begin{tabular}{c|p{3.2cm}|p{10cm}}
    % \hline
    \specialrule{.3em}{.2em}{.2em}
       \multirow{8}{*}{\textbf{Turn 1}} & Original user turn generation & {\textbf{User}(\textcolor{red}{[hotel] area is south , stay is 5 , people is 4}): \textcolor{darkblue}{i need a hotel in the south side please .}}
 \\
 \cline{2-3}
        & \multirow{1}{*}{GPT-3 generated belief state} & 
        <hotel, area, south> , <hotel, book stay, 5> , <hotel, book people, 4>
 \\
        \cline{2-3}
        & \multirow{1}{*}{Revised belief state} & 
        <hotel, area, south> , \sout{<hotel, book stay, 5>} , \sout{<hotel, book people, 4>}, <hotel, type, hotel>
 \\
        \cline{2-3}
        & Revised user turn generation  & {\textbf{User}(\textcolor{red}{[hotel] area is south , type is hotel}): \textcolor{darkblue}{i need a hotel in the south side please .}}
    \\
    \cline{2-3}
        & DB query result  & {[db\_1]}
    \\
    \cline{2-3}
        & Revised dialog act  & <hotel, inform, area>, <hotel, inform, name>, <hotel, inform, internet>, <hotel, inform, parking>, <hotel, inform, type>, <hotel, inform, stars>, <hotel, offerbook, none>
    \\
    \cline{2-3}
        & Revised system turn generation  & {\textbf{Assistant}(\textcolor{red}{[hotel] [inform] area name internet price parking type stars [offerbook]}): \textcolor{darkgreen}{the [value_name] hotel is in the south side and it has [value_internet] . the price is [value_price] per night and it has [value_parking] . it is a [value_stars] star hotel .}}
    \\
    % \hline
    \specialrule{.3em}{.2em}{.2em}
     \multirow{8}{*}{\textbf{Turn 2}} & Original user turn generation & {\textbf{User}(\textcolor{red}{[hotel] stay is 5 , people is 4}): \textcolor{darkblue}{i would like to to book it for 4 people and 5 nights.}}
 \\
 \cline{2-3}
        & \multirow{1}{*}{GPT-3 generated belief state} & 
        <hotel, book stay, 5> , <hotel, book people, 4>
 \\
        \cline{2-3}
        & \multirow{1}{*}{Revised belief state} & 
        <hotel, book stay, 5> , <hotel, book people, 4>
 \\
        \cline{2-3}
        & Revised user turn generation  & {\textbf{User}(\textcolor{red}{[hotel] stay is 5 , people is 4}): \textcolor{darkblue}{i would like to to book it for 4 people and 5 nights.}}
    \\
    \cline{2-3}
        & DB query result  & {[db\_1]}
    \\
    \cline{2-3}
        & Revised dialog act  & <hotel, offerbooked, reference>, <general, reqmore, none>
    \\
    \cline{2-3}
        & Revised system turn generation  & {\textbf{Assistant}(\textcolor{red}{[hotel] [offerbooked] reference [general] [reqmore]}): \textcolor{darkgreen}{your booking reference number is [value_reference] .}}
    \\
    \specialrule{.3em}{.2em}{.2em}
    \multirow{8}{*}{\textbf{Turn 3}} & Original user turn generation & {\textbf{User}(\textcolor{red}{[train] destination is birmingham new street , arrive is 13:06}): \textcolor{darkblue}{i need a train to birmingham new street station that arrives by 13:06 please .}}
 \\
 \cline{2-3}
        & \multirow{1}{*}{GPT-3 generated belief state} & 
        <train, destination, birmingham new street> , <train, arrive, 13:06>
 \\
        \cline{2-3}
        & \multirow{1}{*}{Revised belief state} & 
        <train, destination, birmingham new street> , <train, arrive, 13:06>
 \\
        \cline{2-3}
        & Revised user turn generation  & {\textbf{User}(\textcolor{red}{[train] destination is birmingham new street , arrive is 13:06}): \textcolor{darkblue}{i need a train to birmingham new street station that arrives by 13:06 please .}}
    \\
    \cline{2-3}
        & DB query result  & {[db\_3]}
    \\
    \cline{2-3}
        & Revised dialog act  & <train, request, day>, <train, request, departure>
    \\
    \cline{2-3}
        & Revised system turn generation  & {\textbf{Assistant}(\textcolor{red}{[train] [request] day departure}): \textcolor{darkgreen}{what day will you be leaving and what is your departure station ?}}
    \\
    \specialrule{.3em}{.2em}{.2em}
    \multirow{8}{*}{\textbf{Turn 4}} & Original user turn generation & {\textbf{User}(\textcolor{red}{[train] day is saturday , departure is cambridge}): \textcolor{darkblue}{i will be leaving this saturday from cambridge station .}}
 \\
 \cline{2-3}
        & \multirow{1}{*}{GPT-3 generated belief state} & 
        <train, day, saturday> , <train, departure, cambridge>
 \\
        \cline{2-3}
        & \multirow{1}{*}{Revised belief state} & 
        <train, day, saturday> , <train, departure, cambridge>
 \\
        \cline{2-3}
        & Revised user turn generation  & {\textbf{User}(\textcolor{red}{[train] day is saturday , departure is cambridge}): \textcolor{darkblue}{i will be leaving this saturday from cambridge station .}}
    \\
    \cline{2-3}
        & DB query result  & {[db\_3]}
    \\
    \cline{2-3}
        & Revised dialog act  & <train, inform, arrive>, <train, inform, id>, <train, inform, leave>, <train, offerbook, none>
    \\
    \cline{2-3}
        & Revised system turn generation  & {\textbf{Assistant}(\textcolor{red}{[train] [inform] arrive id leave [offerbook]}): \textcolor{darkgreen}{the train arrives at [value_arrive] and the id is [value_id] . would you like me to book it for you ?}}
    \\
    % \hline
    \specialrule{.3em}{.2em}{.2em}
    \multirow{8}{*}{\textbf{Turn 5}} & Original user turn generation & {\textbf{User}(\textcolor{red}{[train]}): \textcolor{darkblue}{no thank you . what is the cost of the ticket ?}}
 \\
 \cline{2-3}
        & \multirow{1}{*}{GPT-3 generated belief state} & 
        <train, none, none>
 \\
        \cline{2-3}
        & \multirow{1}{*}{Revised belief state} & 
        <train, none, none>
 \\
        \cline{2-3}
        & Revised user turn generation  & {\textbf{User}(\textcolor{red}{[train]}): \textcolor{darkblue}{no thank you . what is the cost of the ticket ?}}
    \\
    \cline{2-3}
        & DB query result  & {[db\_3]}
    \\
    \cline{2-3}
        & Revised dialog act  & <train, inform, price>, <general, reqmore, none>
    \\
    \cline{2-3}
        & Revised system turn generation  & {\textbf{Assistant}(\textcolor{red}{[train] [inform] price [general] [reqmore]}): \textcolor{darkgreen}{the ticket price is [value_price] . can i be of further assistance ?}}
    \\
    \specialrule{.3em}{.2em}{.2em}
    \multirow{8}{*}{\textbf{Turn 6}} & Original user turn generation & {\textbf{User}(\textcolor{red}{[general]}): \textcolor{darkblue}{that is all for now . thanks}}
 \\
 \cline{2-3}
        & \multirow{1}{*}{GPT-3 generated belief state} & 
        <general, none, none>
 \\
        \cline{2-3}
        & \multirow{1}{*}{Revised belief state} & 
        <general, none, none>
 \\
        \cline{2-3}
        & Revised user turn generation  & {\textbf{User}(\textcolor{red}{[general]}): \textcolor{darkblue}{that is all for now . thanks}}
    \\
    \cline{2-3}
        & DB query result  & {[db\_nores]}
    \\
    \cline{2-3}
        & Revised dialog act  & <general, bye, none>
    \\
    \cline{2-3}
        & Revised system turn generation  & {\textbf{Assistant}(\textcolor{red}{[general] [bye]}): \textcolor{darkgreen}{you are welcome , please contact us if you need anything else .}}
    \\
    \specialrule{.3em}{.2em}{.2em}
    \end{tabular}
    % }
    \vspace{1mm}
    \caption{{The controllable dialogue generation process of a dialogue given the prompt in Table~\ref{tab:whole_prompt}. For each turn, GPT-3 will first generate the \textcolor{red}{belief state} and the \textcolor{darkblue}{user utterance}. We parse the belief state, which is then verified and revised automatically. We replace the original generated belief state with the revised belief state in the user turn generation. Then the revised user turn generation will be used to query the database and concatenated with the dialogue context to continue the generation. As for the system turn, we use the revised dialog act, conditioned on which we prompt GPT-3 to generate the system response. Note that the user goal of the target dialogue is allowed to change during the generation. We will keep the updated user goal instead of the original one in the prompt (\textit{instruction3}), which is only used to initiate the generation of the target dialogue.} 
    }
    \label{tab:whole_generation_process}
\end{table*}

\subsection{Turn-level Generation Example}\label{sect:dst_example}
We provide the illustration of three different types of augmented turns for DST in Table~\ref{tab:dst_example}, and the prompt and generation process of it in Table~\ref{tab:dst_prompt}. 

\begin{table*}[htb!]
    \centering
    \footnotesize
    \renewcommand\arraystretch{1.5}
        % \resizebox{\textwidth}{70mm}{
    \begin{tabular}{c|p{12cm}}
    % \hline
    \specialrule{.3em}{.2em}{.2em}
        \multicolumn{2}{c}{Example 1 (\textit{request})} \\
        \hline
        \tabincell{c}{\textbf{Context (SNG01856,}\\\textbf{turn 1)}} & \tabincell{l}{
        \textbf{User}: \textcolor{darkblue}{am looking for a place to to stay that has cheap price range it should be in a type of hotel}\\
        \textbf{Assistant}: \textcolor{darkgreen}{i would be happy to help with your request , what day will you be leaving ?} 
        }
\\
        \hline
        \tabincell{c}{\textbf{Last dialog act}\\\textbf{(turn 1)}} & \tabincell{l}{
        <hotel, \textcolor{red}{request}, area> \\
        }
 \\
        \hline
        \tabincell{c}{\textbf{Original user turn}\\\textbf{(turn 2)}}& \tabincell{l}{
        \textbf{Belief state}: <hotel, parking, yes>, <hotel, pricerange, cheap> \\
        \textbf{User}: \textcolor{darkblue}{no , i just need to make sure it is cheap . oh , and i need parking}\\
        }
\\
        \hline
        \hline
        \tabincell{c}{\textbf{Augmented user turn}\\\textbf{(turn 2)}}& \tabincell{l}{        
        \textbf{Belief state}: <hotel, star, 1>, <hotel, area, west> \\
        \textbf{User}: \textcolor{darkblue}{a 1 star hotel in the west . }\\
        }
\\
    \specialrule{.3em}{.2em}{.2em}
            \multicolumn{2}{c}{Example 2 (\textit{others})} \\
        \hline
        \tabincell{c}{\textbf{Context (SNG01856,}\\\textbf{turn 1-2)}} & \tabincell{l}{
        \textbf{User}: \textcolor{darkblue}{am looking for a place to to stay that has cheap price range it should be in a type of hotel}\\
        \textbf{Assistant}: \textcolor{darkgreen}{i would be happy to help with your request , what day will you be leaving ?} \\
        \textbf{User}: \textcolor{darkblue}{no , i just need to make sure it is cheap . oh , and i need parking}\\
        \textbf{Assistant}: \textcolor{darkgreen}{i found [value_choice] [value_price] [value_type] for you that include -s parking .}\\ \textcolor{darkgreen}{do you like me to book it ?}\\
        }
\\
        \hline
        \tabincell{c}{\textbf{Last dialog act}\\\textbf{(turn 2)}} & \tabincell{l}{
        <hotel, inform, price>, <hotel, inform, choice>, <hotel, inform, parking>, <hotel, inform, type>,  \\
        <hotel, offerbook, none>\\
        }
 \\
        \hline
        \tabincell{c}{\textbf{Original user turn}\\\textbf{(turn 3)}}& \tabincell{l}{
        \textbf{Belief state}: <hotel, book stay, 3>, <hotel, book day, tuesday>, <hotel, book people, 6> \\
        \textbf{User}: \textcolor{darkblue}{yes , please . 6 people 3 nights starting on tuesday .}\\
        }
\\
        \hline
        \hline
        \tabincell{c}{\textbf{Augmented user turn}\\\textbf{(turn 3)}}& \tabincell{l}{        
        \textbf{Belief state}: <hote, book people, 8>, <hotel, stars, 3>, <hotel, book stay, 2>, \\ <hotel, book day, saturday>\\
        \textbf{User}: \textcolor{darkblue}{please book me a room for 8 people on saturday . we will be staying for 2 nights} \\
        \textcolor{darkblue}{ and would like a 3-star hotel .}\\
        }
\\
    \specialrule{.3em}{.2em}{.2em}
            \multicolumn{2}{c}{Example 3 (\textit{reqmore})} \\
        \hline
            \tabincell{c}{\textbf{Context (SNG01856,}\\\textbf{turn 1-3)}} & \tabincell{l}{
        \textbf{User}: \textcolor{darkblue}{am looking for a place to to stay that has cheap price range it should be in a type of hotel}\\
        \textbf{Assistant}: \textcolor{darkgreen}{i would be happy to help with your request , what day will you be leaving ?} \\
        \textbf{User}: \textcolor{darkblue}{no , i just need to make sure it is cheap . oh , and i need parking}\\
        \textbf{Assistant}: \textcolor{darkgreen}{i found [value_choice] [value_price] [value_type] for you that include -s parking . } \\ \textcolor{darkgreen}{do you like me to book it ?}\\
        \textbf{User}: \textcolor{darkblue}{how about only 2 nights .}\\
        \textbf{Assistant}: \textcolor{darkgreen}{booking was successful . reference number is : [value_reference] . anything else}\\ \textcolor{darkgreen}{ i can do for you ?}\\
        }
\\
        \hline
        \tabincell{c}{\textbf{Last dialog act}\\\textbf{(turn 3)}} & \tabincell{l}{
        <hotel, offerbooked, reference>, <general, \textcolor{red}{reqmore}, none>\\
        }
 \\
        \hline
        \tabincell{c}{\textbf{Original user turn}\\\textbf{(turn 4)}}& \tabincell{l}{
        \textbf{Belief state}: <general, none, none> \\
        \textbf{User}: \textcolor{darkblue}{no , that will be all . goodbye .}\\
        }
\\
        \hline
        \hline
        \tabincell{c}{\textbf{Augmented user turn}\\\textbf{(turn 4)}}& \tabincell{l}{        
        \textbf{Belief state}: <taxi, destination, avalon>\\
        \textbf{User}: \textcolor{darkblue}{i need a taxi to avalon .}\\
        }
\\
    \specialrule{.3em}{.2em}{.2em}
    \end{tabular}
    % }
    \vspace{1.5mm}
    \caption{Example of three augmented user turns for DST in \textbf{SNG01856} (MultiWOZ). These three turns are augmented according to different types of dialog acts in the last turn, which are \textit{request}, \textit{others}, and \textit{reqmore}, respectively. We show the prompt and generation of Example 2 in Table~\ref{tab:dst_prompt}.}
    \label{tab:dst_example}
\end{table*}

\begin{table*}[htb!]
    \centering
    \footnotesize
    \renewcommand\arraystretch{1.5}
        % \resizebox{\textwidth}{70mm}{
    \begin{tabular}{c|p{13cm}}
    % \hline
    \specialrule{.3em}{.2em}{.2em}
        \tabincell{c}{\textbf{Task}\\\textbf{Description}} & \tabincell{l}{Answer the assistant's question on each feature you require when booking a train. Also mention  \\no preference on a feature when your requirement on it is "dontcare". \\
        \textbf{Features}:\\
        people: number of people for the hotel booking;\\
        type: what is the type of the hotel, guesthouse, guest house, or hotel;\\
        stay: length of stay at the hotel;\\
        name: name of the hotel;\\
        day: day of the hotel booking;\\
        stars: star rating of the hotel;\\
        }
 \\
        \hline
        \tabincell{c}{\textbf{Example 1}\\\textbf{(PMUL1576,}\\\textbf{turn 3)}} & \tabincell{l}{\textbf{Assistant}: what is your requirement on day? \\
\textbf{User}(\textcolor{red}{[hotel] day is friday}): \textcolor{darkblue}{yes , please book me a room for friday .}}
 \\       
        \hline
        \tabincell{c}{\textbf{Example 2}\\\textbf{(SNG1006,}\\\textbf{turn 2)}} & \tabincell{l}{\textbf{Assistant}: what is your requirement on type, name, stay, day, people? \\
\textbf{User}(\textcolor{red}{[hotel] type is hotel , name is gonville hotel , stay is 4 , day is saturday , people is 6}): \\
\textcolor{darkblue}{okay , i would like to book a room at the gonville hotel for 4 nights . there will be 6 people and }\\ \textcolor{darkblue}{we will be arriving on saturday .}}
 \\
 
        \hline
        \tabincell{c}{\textbf{Target}\\\textbf{Turn}\\\textbf{Prompt}} & \tabincell{l}{\textbf{User}(\textcolor{red}{[hotel] people is 8 , stars is 3 , stay is 2 , day is tuesday}):}
\\
        \hline
        \tabincell{c}{\textbf{Target}\\\textbf{Turn}\\\textbf{Completion}} & \tabincell{l}{\textbf{User}(\textcolor{red}{[hotel] people is 8 , stars is 3 , stay is 2 , day is tuesday}): \\                                                    
        \textcolor{darkblue}{please book me a room for 8 people on tuesday . we will be staying for 2 nights and would like a}\\ \textcolor{darkblue}{3-star hotel .}}
\\
    % \hline
    \specialrule{.3em}{.2em}{.2em}
    \end{tabular}
    % }
    \vspace{1.5mm}
    \caption{Example of a complete prompt for the generation of Example 2 in Table~\ref{tab:dst_example}. The task description contains the description of the mentioned slots to help GPT-3 better understand. We sample some turns from seed dialogues instead of the whole dialogues. Given the target turn-level \textcolor{red}{belief state}, GPT-3 is able to generate the \textcolor{darkblue}{user utterance} that expresses the user goal. }
    \label{tab:dst_prompt}
\end{table*}

\end{document}